\documentclass[11pt,a4paper]{article}
\usepackage[hyperref]{acl2021}
\usepackage{times}
\usepackage{latexsym}

\usepackage{multirow}
\usepackage{tabularx}
\usepackage{microtype}
\usepackage{amsthm,amsmath}
\usepackage{graphicx}
\usepackage{mathrsfs}
\usepackage{appendix}
\usepackage{url}
\usepackage{soul}
\usepackage{enumitem}
\usepackage{multirow}
\usepackage{hyperref}
\usepackage{dsfont}
\usepackage{booktabs}
\usepackage{amsfonts}
\usepackage{bm}
\usepackage{todonotes}
\usepackage[ruled,linesnumbered, vlined]{algorithm2e} 

\setlist[itemize]{noitemsep, topsep=0pt,left=0pt}

\newcommand\blfootnote[1]{%
  \begingroup
  \renewcommand\thefootnote{}\footnote{#1}%
  \addtocounter{footnote}{-1}%
  \endgroup
}

\aclfinalcopy 
\def\aclpaperid{1877} 

\makeatletter
\newcommand{\removelatexerror}{\let\@latex@error\@gobble}
\makeatother

\title{Adaptive Knowledge-Enhanced Bayesian Meta-Learning for Few-shot Event Detection}

\author{Shirong Shen$^{1}$ \and
Tongtong Wu$^{1}$ \and
Guilin Qi$^{1*}$ \and
Yuan-Fang Li$^{2}$ \and \\
\and \textbf{Gholamreza Haffari} $^{2}$
 \textbf{Sheng Bi}$^{1}$
 \\
$^{1}$School of Computer Science and Engineering, Southeast University, China \\
$^{2}$Faculty of Information Technology, Monash University, Melbourne, Australia \\
\tt\{ssr, wutong8023, gqi\}@seu.edu.cn,  yuanfang.li@monash.edu, \\ \tt  Gholamreza.Haffari@monash.edu, bisheng@seu.edu.cn}

\date{}

\begin{document}

\maketitle
\begin{abstract}
Event detection (ED) aims at detecting event trigger words in sentences and classifying them into specific event types. In real-world applications, ED typically does not have sufficient labelled data, thus can be formulated as a few-shot learning problem. To tackle the issue of low sample diversity in few-shot ED, we propose a novel knowledge-based few-shot event detection method which uses a definition-based encoder to introduce external event knowledge as the knowledge prior of event types. Furthermore, as external knowledge typically provides limited and imperfect coverage of event types, we  introduce an adaptive knowledge-enhanced Bayesian meta-learning method to dynamically adjust the knowledge prior of event types. Experiments show  our method consistently and substantially outperforms a number of baselines by at least 15 absolute $F_1$ points under the same few-shot settings. 
\end{abstract}
\blfootnote{
  \hspace{-0.2cm}
  * Corresponding author.}


\section{Introduction}
\label{sec:intro}
Event detection is an important task in information extraction, aiming at detecting event triggers from text and then classifying them into event types~ \cite{chen2015event}. For example, in ``\emph{The police \textbf{arrested} Harry on charges of manslaughter}'', 
the trigger word is \emph{arrested}, indicating an \textbf{``Arrest''} event.
Event detection has been widely applied in Twitter analysis~\cite{twitter-17}, legal case extraction~\cite{legal_17}, and financial event extraction~\cite{financial_19}, to name a few.

Typical approaches to event detection~\cite{chen2015event,mcclosky2011event,liu2019event} generally rely on large-scale annotated datasets for training. Yet in real-world applications, adequate labeled data is usually unavailable.
%
Hence, methods that generalize effectively with small quantities of labeled samples and adapt quickly to new event types are highly desirable for event detection. 

\begin{figure}[t]
    \centering
    \resizebox{7cm}{!}{
    \includegraphics{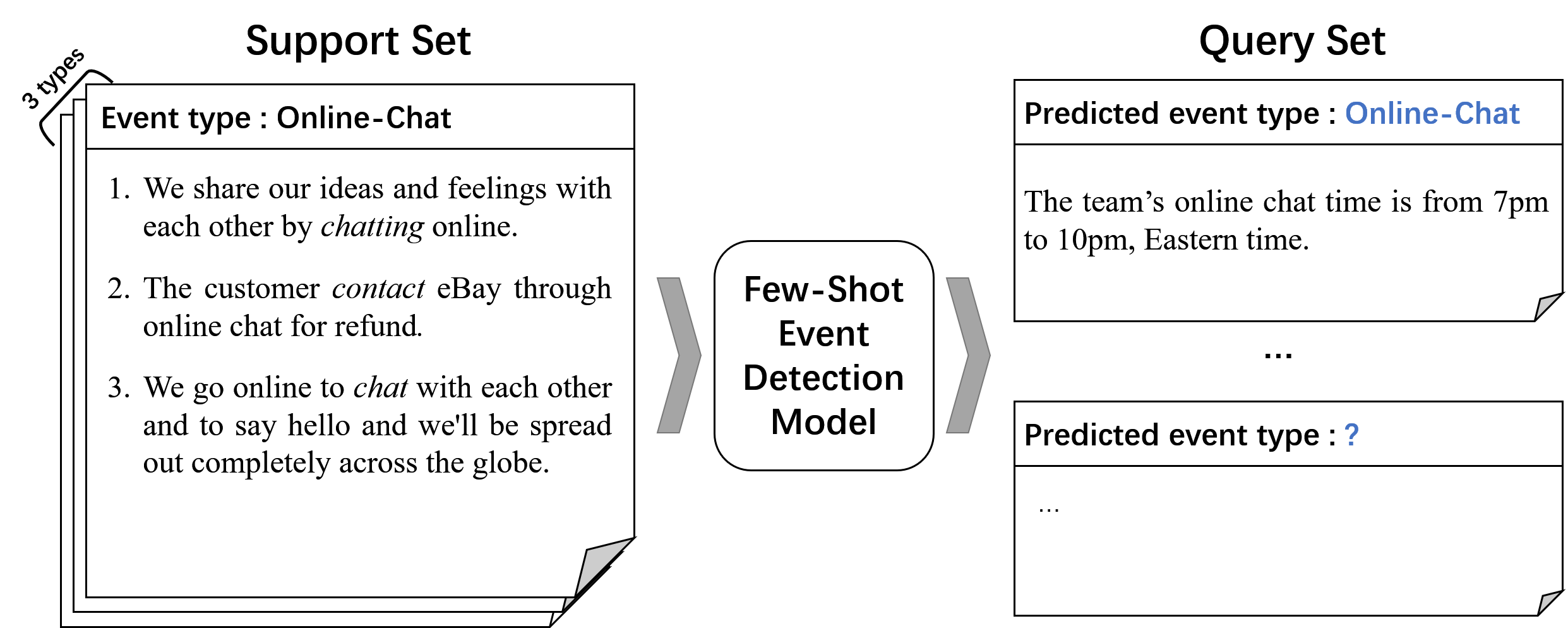}}
    \caption{A 3-way 3-shot event detection example, in which the model uses the support set to predict the event types of query samples.}
    \label{fig:example}
    \vspace{-3mm}
\end{figure}

Various approaches have been proposed to enable learning from only a few samples, i.e., few-shot learning~\cite{finn2017model,snell2017prototypical,zhang18metagan_model}. 
Yet few-shot event detection (FSED) has  been less studied until recently~\cite{lai2020exploiting,deng2020meta}. 
Although these methods achieve encouraging progress on typical $N$-way $M$-shot setting (Figure~\ref{fig:example}), the performance remains unsatisfactory as the \emph{diversity} of examples in the support set is usually limited.

\begin{figure}[t]
    \centering
    \resizebox{7.8cm}{!}{
    \includegraphics{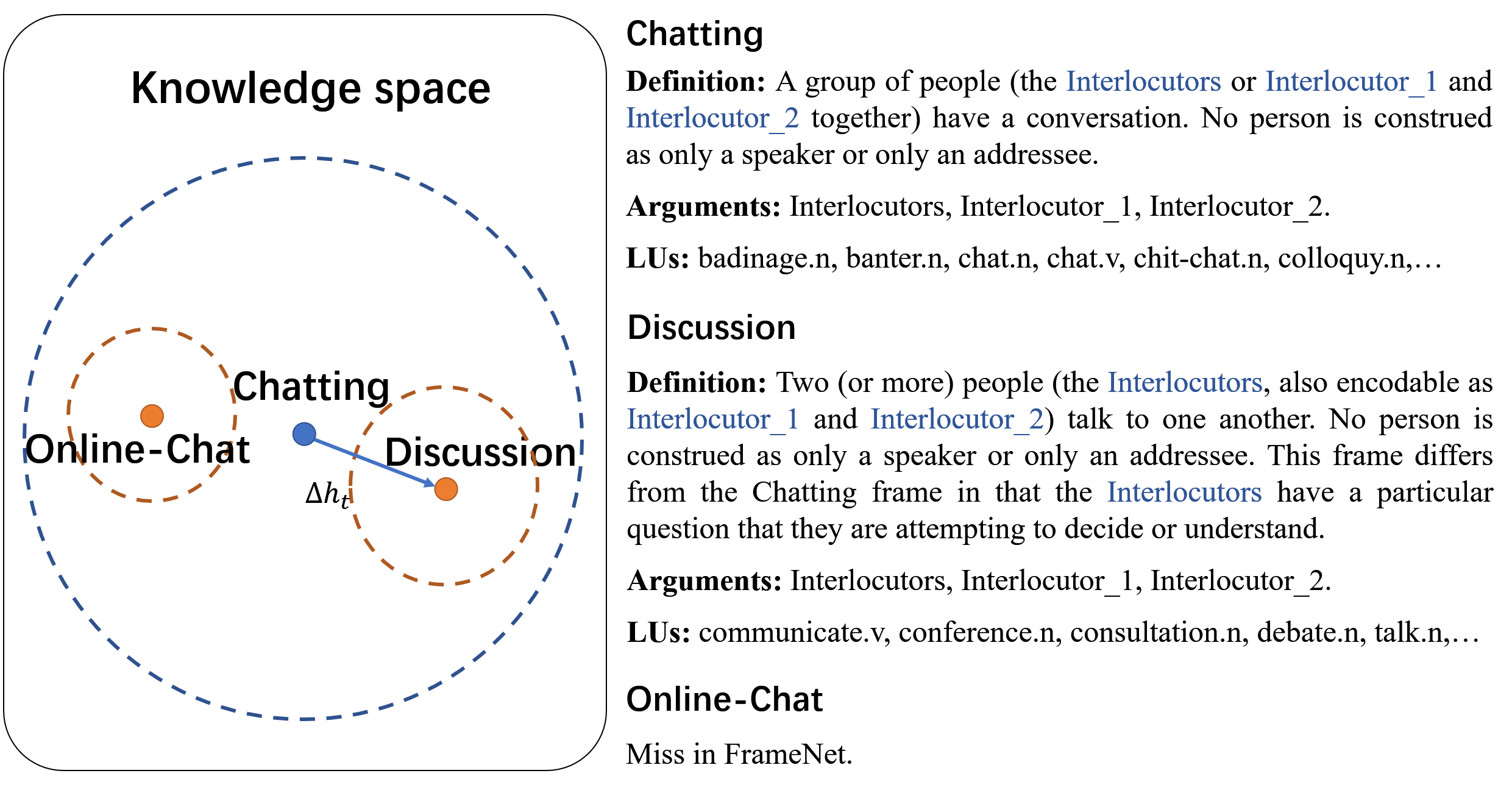}}
    \caption{An example of FrameNet. Left side: the relation between frame ‘Chatting’ and its sub-frame. Right side: the definitions and LUs (Lexical Units) of frame Chatting and Discussion. The blue words represent the mentions of arguments in definition. It can be seen that, in FrameNet, the definition of the sub-frame is similar to the definition of its super-frame. External knowledge base can provide rich semantic information, yet the knowledge base is typically incomplete, such as the missing of a desired frame ``online-chat''.}
    \label{fig:example1}
    \vspace{-4mm}
\end{figure}

Intuitively, introducing high-quality semantic knowledge, such as FrameNet~\cite{baker1998berkeley}, is a potential solution to the insufficient diversity issue~\cite{qu2020few,TongXWCHLX20detection,liu2016leveraging,Liu0020application}.
%
%
However, as shown in Figure~\ref{fig:example1}, such knowledge-enhanced methods also suffer from two major issues: (1) the incomplete coverage  by the knowledge base and (2) the uncertainty caused by the inexact alignment between predefined knowledge and diverse applications. 

To tackle the above issues, in this paper, we propose an Adaptive Knowledge-Enhanced Bayesian Meta-Learning (AKE-BML) framework. More specifically, (1) we align the event types between the support set and FrameNet via heuristic rules.\footnote
{For event types that cannot be accurately aligned to FrameNet, we match the nearest super-ordinate frame for them.}
(2) {We propose encoders for encoding the samples and knowledge-base in the same semantic space}.
(3) {We propose a learnable offset for revising the aligned knowledge representations to build the knowledge prior distribution for event types and generate the posterior distribution for event type prototype representations}. 
(4) In the prediction phrase, we adopt the learned {posterior distribution for prototype representations} to classify query instances into event types. 

We conduct comprehensive experiments on the aggregated benchmark dataset of few-shot event detection~\cite{deng2020meta}. The experimental results show that our method consistently and substantially outperforms state-of-the-art methods. In all six $N$-way-$M$-shot settings, our model achieves a large $F_1$ superiority of at least 15 absolute points.

\section{Related Work}
\paragraph{Event Detection.}
Recent event detection methods based on neural networks have achieved good performance~\cite{chen2015event,sha2016rbpb,nguyen2016joint,ACL2021_MLBiNet}. 
These methods use neural networks to construct the context features of candidate trigger words to classify events.
Pre-trained language models such as BERT~\cite{devlin2018bert} have also become an indispensable component of event detection models~\cite{yang2019exploring,wadden2019entity,shen2020hierarchical}.
However, neural models rely on large-scale labeled event datasets and fail to predict the labels of new event types.
A recent study utilized the basic metric-based few-shot learning method for event detection~\cite{lai2020extensively}.
Deng et al.~\shortcite{deng2020meta} tackles few-shot learning for event classification with a dynamic memory network.
To enhance background knowledge, ontology embedding is used in ED \cite{ACL2021_OntoED}.
These methods have achieved encouraging results in the few-shot learning setting.
However, they do not address the problem of insufficient sample diversity in the support set.
Our method leverages the knowledge in \emph{FrameNet} to augment the support set for event detection.

\paragraph{Few-shot Learning and Meta-learning.}
Few-shot learning trains a model with only a few labeled samples in a support set and predicts the labels of unlabeled samples in the query set.
Various approaches have been proposed to solve the few-shot learning problem, which mainly fall into three categories: (1) metric-based methods~\cite{vinyals2016matching,snell2017prototypical,garcia2017few,sung2018learning}, 
%
(2) optimization-based methods~\cite{finn2017model,nichol18reptile,ravi2016optimization}, and 
%
(3) model-based methods~\cite{yan15model,zhang18model,sukh15memory_model,zhang18metagan_model}.
%
However, these methods rely heavily on the support set and suffer from poor robustness caused by insufficient sample diversity of the support set.

Bayesian meta-learning~\cite{ravi2016optimization,yoon2018bayesian} can construct the posterior distribution of the prototype vector through external information outside the support set. The effectiveness of this method has been shown in the few-shot relation extraction task~\cite{qu2020few}. 
It inspires us to solve the problem of insufficient sample diversity in the task of few-shot event detection by introducing external knowledge. 
However, this method ignores the semantic deviation between knowledge and target types.
Specifically, a knowledge base may provide incomplete coverage of target types in a given support set, which leads to inaccurate matching between a target type and knowledge.

\section{Problem Definition}\label{sec:probdef}

In this paper, the Few-Shot Event Detection (FSED) problem is defined as a typical \emph{N-way-M-shot} problem. 
Specifically, a tiny labeled \emph{support set} $S$ is provided for model training. $S$ contains \emph{N} distinct event types and each event type has only \emph{M} labeled samples, where \emph{M} is typically small (e.g. $M=5, 10, 15$).
%
%
%
More precisely, in each FSED task we are given a small support set $S=\{(x_s,y_s)\}$. Let $X_S=\{x_s\}_{s \in S}$ represent the samples in the support set $S$, i.e.\ $x_s=(I_s,tt_s)$, where $I_s$ is the sentence of the sample $x_s$ and $tt_s$ is the candidate trigger word of $x_s$. 
We denote by $Y_S$ an ordered list of event types, i.e.\ $Y_S=\{y_s\}_{s\in S}$, where each $y_s$ is the ground-truth event type of sample $x_s$. 
For each support set $S$, we only consider a subset of event types $T_s$ from the entire set of event types $T$.
Hence, in the $N$-way-$M$-shot setting, $|T_S|=N$ and $|X_S|=|Y_S|=N*M$. 

Moreover, we assume an external knowledge base $\mathscr{F}$ that contains a number of \emph{frames}. Each frame $F_t\in\mathscr{F}$ consists of three parts: $F_t = (D_t, A_t, L_t)$, where $D_t$, $A_t$ and $L_t$ are the definition, arguments, and linguistic units (LUs) of the frame respectively. Please see Appendix~\ref{sec:knowledge} for details of FrameNet.

For each support set $S$, we are also given a query set $Q$ composed of some unlabeled samples $X_Q=\{x_q\}_{q \in Q}$, where $x_q=(I_q,tt_q)$, $I_q$ is the sentence of sample $x_q$, and $tt_q$ is the candidate trigger word of $x_q$. 
Our goal is to learn a neural classifier for these event types by using the external knowledge and the support set. 
We will apply the classifier to predict the labels of the query samples in $Q$, i.e., $Y_Q=\{y_q \}_{q\in Q}$ with each $y_q\in T_S$.
We do this by learning 
 $p(Y_Q |X_Q,X_S,Y_S,\mathscr{F})$.

%

\section{Adaptive Knowledge-Enhanced Bayesian Meta-Learning}


\begin{figure*}
    \centering
    \resizebox{13cm}{!}{
    \includegraphics{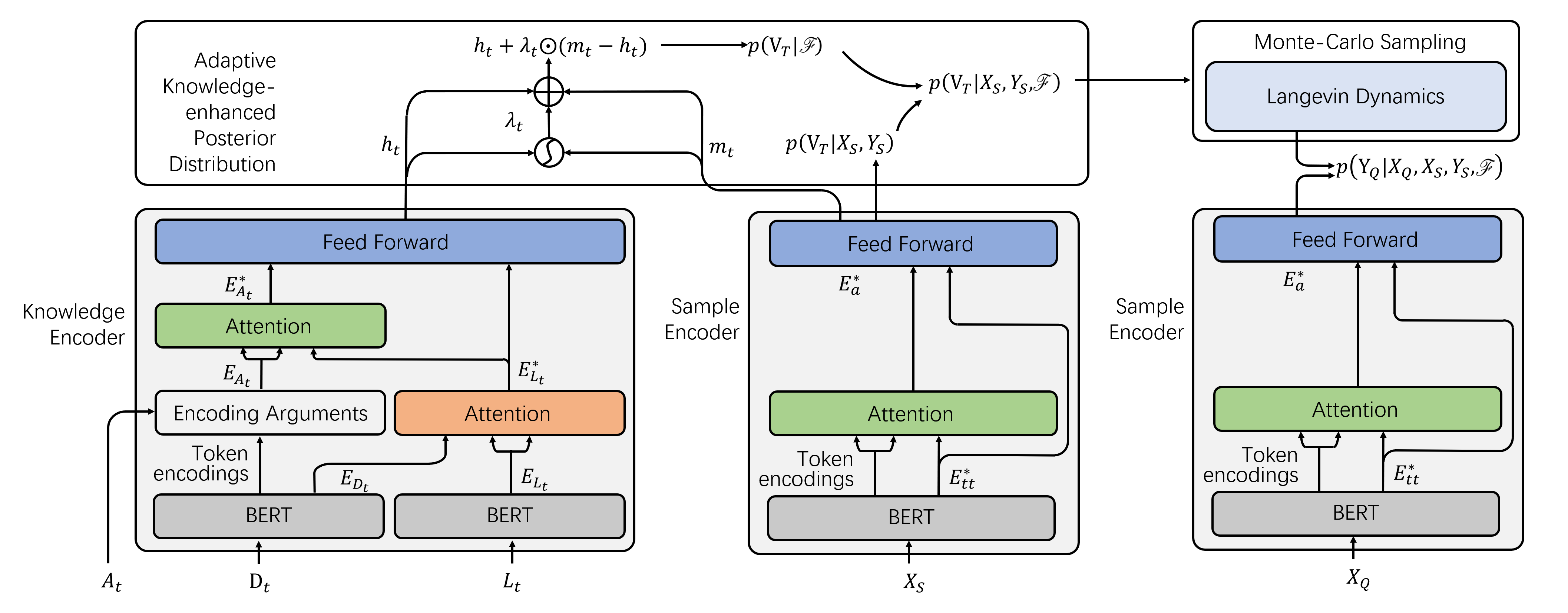}}
    \caption{Framework overview. Our method combines both the external knowledge and the support set into a prior distribution of event prototype.  We customize two encoders to generate sample representations and knowledge representations. Then we utilize the support set to generate a learnable offset for revising the aligned knowledge representations to generate the prior distribution for prototype representations. Finally, we use Monte-Carlo sampling and stochastic gradient Langevin dynamics to draw samples of prototypes for prediction.}
    \label{fig:over}
\end{figure*}

We now present our adaptive knowledge-enhanced few-shot event detection approach.
The overall structure of our method is shown in Figure \ref{fig:over}.
Our method represents each event type $t$ with a prototype vector $\textbf{v}_t$, which is then used to classify the query sentences.
We use $\textbf{V}_{T_S} = \{ \textbf{v}_t \}_{t \in T_S}$ to represent the collection of prototype vectors for all event types in $T_S$.
Then the conditional  distribution $p(Y_Q |X_Q,X_S,Y_S,\mathscr{F})$ can be represented as:
\vspace{-3mm}
\begin{equation}
    \label{eq:divide}
    \int p(Y_Q |X_Q,V_{T_S})p(V_{T_S}|X_S,Y_S,\mathscr{F}) \, d\textbf{V}_{T_S}.
\end{equation}
To calculate Eq. \ref{eq:divide}, we first introduce the sample encoder and knowledge encoder to give the vector representations of samples and the knowledge of event types.
Then we use the sample representations and knowledge representation to construct the adaptive knowledge-enhanced posterior distribution $p(V_{T_S}|X_S,Y_S,\mathscr{F})$ of $\textbf{V}_{T_S}$ and give the likelihood $p(Y_Q |X_Q,V_{T_S})$ by $\textbf{V}_{T_S}$ and sample representations.
Finally we leverage Monte Carlo sampling to approximate the posterior distribution and draw each prototype sample by the stochastic gradient Langevin dynamics~\cite{welling2011bayesian} to optimize model parameters in an end-to-end fashion.
We now explain the framework in more details.

\subsection{Sample and Knowledge Encoder}

The purpose of encoding knowledge is to make up for the lack of diversity and coverage of the support set.  Thus we align the knowledge and sample encoding and map them into the same semantic space. 
Intuitively, trigger and arguments are the main factors for entity detection. Hence, to align the trigger and arguments from samples and external knowledge, we design two encoders for the knowledge and samples, generating the final knowledge encoding $\textbf{h}_t$ and the sample encoding $\bm{\mathcal{E}}(x)$ with the same dimensions.

\noindent \textbf{Knowledge Encoder.}
Given a knowledge frame $F_t = \{D_t, A_t, L_t\}$ for the event type $t$, 
we encode it into a real-valued vector to represent the semantics of $t$. As shown in Figure \ref{fig:example1}, for a frame $F_t$, the linguistic units $L_t$ can represent the features of the trigger words, the arguments $A_t$ can represent the context of the trigger words in   samples, and $D_t$ describes the semantic relationship between $A_t$ and $t$. 

For each event type $t$, the proposed knowledge encoder uses BERT to generate the text encoding $\textbf{E}_{D_t}$ and $\textbf{E}_{L_t}$ from the description $D_t$ and the LUs $L_t$ respectively. 
Moreover, the arguments encoding $\textbf{E}_{A_t}$ is a sequence of $\textbf{e}_{A_t}^{(i)}$, i.e., the average token encoding in the $i$-th argument mention in $D_t$, which ensures that the encoding of $A_t$ fully contains the semantics of the event type $t$.
Then, as shown in Figure~\ref{fig:over}, the trigger word prior encoding  and argument prior encoding are generated by follows:

\begin{itemize}
    \item \textbf{Trigger word prior encoding.} We use  attention  to get the weighted sum of words in $L_t$ as the trigger word prior encoding $\textbf{e}^*_{L_t}$. The \emph{query} of the attention is $\textbf{E}_{D_t}$, \emph{key} and \emph{value} are both $\textbf{E}_{L_t}$.
    \item \textbf{Argument prior encoding.} 
    An attention mechanism is used to aggregate the arguments information into $\textbf{e}^*_{A_t}$, where the \emph{query} of the attention is $\textbf{e}^*_{L_t}$, \emph{key} and \emph{value} are both $\textbf{E}_{A_t}$.
\end{itemize}

Finally, we concatenate the trigger word prior encoding $\textbf{e}^*_{L_t}$ and the argument prior encoding $\textbf{e}^*_{A_t}$, and use a feed forward network $f^h$ to generate the knowledge encoding vector $\textbf{h}_t$ of event type $t$,
\begin{equation}
\label{eq:kencoder}
    \textbf{h}_t = f^h \left( \left[\textbf{e}^*_{A_t};\textbf{e}^*_{L_t} \right] \right).
\end{equation}

\noindent\textbf{Sample encoder.}
We follow the same strategy to build a sample encoder. Given each sample $x=(I,tt)$, i.e., a candidate trigger word $tt$ and its context $I$, we first utilize BERT to encode $x$ and select the encoding of $tt$ as the trigger representation $\textbf{e}^*_{tt}$. As arguments are not explicitly given in $x$, we use an attention mechanism to aggregate the implicit argument information for current trigger $tt$, in which the \emph{query} is $\textbf{e}^*_{tt}$, \emph{key} and \emph{value} are both token encoding generate from $I$. We denote the argument encoding as $\textbf{e}^*_a$.

Finally, we concatenate the trigger word encoding $\textbf{e}^*_{tt}$ and the argument encoding $\textbf{e}^*_{a}$, and use a feed forward network $f^{\mathcal{E}}$ to generate the sample encoding vector $\bm{\mathcal{E}}( x )$,
\begin{equation}
    \label{eq:sencoder}
    \bm{\mathcal{E}}( x ) = f^{\mathcal{E}} \left( \left[\textbf{e}^*_a;\textbf{e}^*_{tt} \right] \right).
\end{equation}

\subsection{Adaptive Knowledge-Enhanced Posterior} 
\label{sec:parapost}

The posterior distribution can be factorized into a prior distribution (given the event knowledge) and a likelihood  on the support set \cite{qu2020few} as, 
\begin{equation}
    \label{eq:postfac}
       p(\textbf{V}_{T_S} |X_S,Y_S,\mathscr{F}) \propto p(Y_S|X_S,\textbf{V}_{T_S})p(\textbf{V}_{T_S}|\mathscr{F}),
\end{equation}
where $p(Y_S|X_S,\textbf{V}_{T_S})$ is the likelihood on the support set, and $p(\textbf{V}_{T_S}|\mathscr{F})$ is the adaptive knowledge-based prior for the prototype vectors. We describe the details of these two components as follows:

\noindent\textbf{Adaptive Knowledge-based Prior.}
As we discussed in Section~\ref{sec:intro}, an event type $t$ may not have an exact/perfect match in the knowledge base $\mathscr{F}$. In such situations, we resort to finding the super-ordinate frame of $t$, which is semantically closest to $t$. As shown in Figures~\ref{fig:example} and~\ref{fig:case}, where the event type $t$ in the support set `online-chat' is matched against the knowledge prior $F_t$ `Chatting' in FrameNet, a super-ordinate frame.
In order to enable the knowledge encoding to accurately reflect the characteristics of the corresponding event type, we add a learnable \emph{knowledge offset} to $\textbf{h}_t$.
We denote the knowledge offset between the event type $t$ and its knowledge encoding $\textbf{h}_t$ by $\Delta \textbf{h}_t$.
Recall that the knowledge in $\textbf{h}_t$ is encoded from the exactly-matched frame or the super-ordinate frame. $\Delta \textbf{h}_t$ is defined as follows:
\vspace{-2mm}
\begin{equation}
    \label{eq:delt}
    \Delta \textbf{h}_t = \bm{\lambda}_{t} \odot (\textbf{m}_t - \textbf{h}_t),
\vspace{-2mm}
\end{equation}
where $\odot$ is the element-wise product, and $\textbf{m}_t$ is the mean  of the encodings $\bm{\mathcal{E}}(x)$ of all the samples $x$ in the support set.
%
 $\bm{\lambda}_t\in [0,1]^{|\textbf{h}_t|}$ is the adaptive weight (gate), which  is obtained from the sample encoding $\textbf{m}_t$ and the knowledge encoding $\textbf{h}_t$:
 \vspace{-2mm}
\begin{equation}
    \label{eq:lambda}
    \bm{\lambda}_{t} = \sigma(\textbf{W}_{\lambda}\left[\textbf{m}_t;\textbf{m}_t-\textbf{h}_t;\textbf{h}_t \right] + \textbf{b}_{\lambda}),
\vspace{-2mm}
\end{equation}
where $\sigma$ is the nonlinear sigmoid  function, 
and $\textbf{W}_{\lambda}$ and $\textbf{b}_{\lambda}$ are trainable parameters. 


%
%

Putting it altogether, the knowledge prior distribution has the following form,
\vspace{-2mm}
\begin{equation}
    \label{eq:priorfixed}
     \begin{aligned}
         p(\textbf{V}_{T_S} |\mathscr{F}) &= \prod_{t \in T_S} p(\textbf{v}_t|\textbf{h}_t,\Delta \textbf{h}_t) \\
       &= \prod_{t \in T_S} \mathcal{N}(\textbf{v}_t|\textbf{h}_t + \Delta \textbf{h}_t,\boldsymbol{I}),
     \end{aligned}
\vspace{-2mm}     
\end{equation}
where $\mathcal{N}(\textbf{v}_t|\textbf{h}_t + \Delta \textbf{h}_t,\boldsymbol{I})$ is multivariate Gaussian  with the mean $\textbf{h}_t + \Delta \textbf{h}_t$ and covariance $\boldsymbol{I}$ (the identity matrix).  
So, each prototype vector has a prior distribution  containing knowledge from \emph{FrameNet} adaptively adjusted according to the support set. 


\noindent\textbf{Likelihood.}
With the given prototype vectors $\textbf{V}_{T_s}$ distributed according to $p(\textbf{V}_{T_S}|X_S,Y_S,\mathscr{F})$, the likelihood for support samples is defined as,
\vspace{-2mm}  
\begin{eqnarray}
\label{eq:supportprob}
 p(Y_S |X_S,\textbf{V}_{T_S}) &=& \prod_{s \in S} p(y_s|x_s,\textbf{V}_{T_S}) \\
p(y_s = t |x_s,\textbf{V}_{T_S}) 
        &:=& \frac{\exp(\bm{\mathcal{E}}(x_s) \cdot \textbf{v}_t)}{\sum_{t' \in T_S} \exp(\bm{\mathcal{E}}(x_s) \cdot \textbf{v}_{t'})}. \nonumber
\vspace{-2mm}  
\end{eqnarray}
The dot product of the sample encoding $\bm{\mathcal{E}}(x_q)$ and the event type prototype vector $\textbf{v}_t$ estimates their similarity. We use $softmax$  to normalize the result to the probability of $x_s$ belonging to event type $t$.

\subsection{Optimization and Prediction}
\label{sec:optpred}
For prediction, the model computes and maximizes the log-probability $\log p(Y_Q |X_Q,X_S,Y_S,\mathscr{F})$. However, according to Eqn~(\ref{eq:divide}), the log-probability relies on the integration over prototype vectors, which is difficult to compute. Hence, we estimate it with Monte Carlo sampling~\cite{qu2020few},
\vspace{-2mm}  
\begin{eqnarray}
    \label{eq:Monte}
    && p(Y_Q |X_Q,X_S,Y_S,\mathscr{F})  \nonumber \\
      &&=\mathbb{E}_{ p(\textbf{V}_{T_S}|X_S,Y_S,\mathscr{F}) } \left[ p(Y_Q |X_Q,\textbf{V}_{T_S}) \right] \nonumber  \\
       &&\approx \frac{1}{N_s} \sum_{i=1}^{N_s} p(Y_Q |X_Q,\textbf{V}_{T_S}^{ \left( s \right)})
\end{eqnarray}
where $N_s$ is the number of samples, and $\textbf{V}_{T_S}^{ \left( s \right)}$ is a sample drawn from the posterior distribution, i.e.\ $\textbf{V}_{T_S}^{ \left( s \right)} \sim p(\textbf{V}_{T_S}|X_S,Y_S,\mathscr{F}) $. 
$p(Y_Q |X_Q,\textbf{V}_{T_S}^{ \left( s \right)})$ is the likelihood for query samples which has the same form as Eqn~\ref{eq:supportprob}.
To sample from the posterior, we  use the stochastic gradient Langevin dynamics~\cite{welling2011bayesian} with multiple stochastic updates. Formally, we  initialize the sample $\hat{\textbf{V}}_{T_S}$ and iteratively update the sample as,
\vspace{-2mm}
\begin{align}
        \hat{\textbf{V}}_{T_S} \gets &\hat{\textbf{V}}_{T_S} + \sqrt{\epsilon} \boldsymbol{z}  \label{eq:update0}\\
        &+ \frac{\epsilon}{2} \nabla_{\hat{\textbf{V}}_{T_S}} \log \ p(Y_S|X_S,\hat{\textbf{V}}_{T_S})p(\hat{\textbf{V}}_{T_S}|\mathscr{F}), \nonumber 
\vspace{-2mm} 
\end{align}
%
where $\boldsymbol{z} \sim \mathcal{N}(\boldsymbol{0},\boldsymbol{I})$, and $\epsilon$ is a small real number representing the update step size. 
The gradient $\nabla_{\hat{\textbf{V}}_{T_S}} \log \ p(Y_S|X_S,\hat{\textbf{V}}_{T_S})p(\hat{\textbf{V}}_{T_S}|\mathscr{F})$ in Eqn~\ref{eq:update0} balances the effect of the knowledge and the support set on the prototype vector.  
Please see Appendix~\ref{A:GPD} for derivation details and intuitive explanations of its influence. 

The Langevin dynamics requires a burn-in period.  To speed up the convergence, we follow the previous method \cite{qu2020few} and initialize the sample as follows, 
\vspace{-2mm} 
\begin{equation}
    \label{eq:initialize}
    \begin{aligned}
        \hat{\textbf{V}}_{T_S} &\gets \{ \hat{\textbf{v}}_{t} \}_{t \in T_S} \\
        \hat{\textbf{v}}_{t} &\gets \textbf{m}_t + \textbf{h}_t + \Delta \textbf{h}_t - \textbf{m},
    \end{aligned}
\vspace{-2mm}     
\end{equation}
where $\textbf{m}$ is the mean encoding of all the samples in the support set. 

After obtaining  prototype samples from the posterior, $\log p(Y_Q |X_Q,X_S,Y_S,\mathscr{F})$ is end-to-end approximated according to Eqn~(\ref{eq:Monte}). 
During the training stage, we optimize the log-likelihood of the query set and update the model parameters by gradient descent. 
In the prediction stage, the log-likelihood will determine the probability that a query sample belongs to each event type. The training process is shown in Algorithm \ref{alg:A}.

\begin{figure}
\small
\removelatexerror
\begin{algorithm}[H]
\caption{Training Process}
\label{alg:A}
\KwIn{Event type set $T$}
\KwIn{Event knowledge from \emph{FrameNet}}

\While {not convergence} {

    Sample a subset $T_S$ from $T$ to build a FSED task 
    
    Sample disjoint support and query sets for $T_S$
    
    Compute the sample encodings (Eq.~\ref{eq:sencoder})
    
    Compute the $\{\textbf{m}_t\}_{t \in T_S}$ for each $t \in T_S$ 
    
    Compute  knowledge encodings $\{\textbf{h}_t\}_{t \in T_S}$ (Eq.~\ref{eq:kencoder})
    
    Compute  knowledge offset $\{\Delta \textbf{h}_t\}_{t \in T_S}$ (Eq.~\ref{eq:delt})
    
    Initialize  prototype vectors $\{\textbf{V}_{T_S}^{s}\}_{s=1}^{N_s}$ (Eq.~ \ref{eq:initialize})
    
    Update prototype vectors iteratively  (Eq. \ref{eq:update})
    
    Compute and maximize log-likelihood (Eq.~\ref{eq:Monte})
}
\end{algorithm}
\vspace{-6mm}
\end{figure}

\section{Experiments}

We conduct evaluation with the following goals: (1) to compare our adaptive knowledge-enhanced Bayesian meta-learning method with existing few-shot event detection methods and few-shot learning baseline methods; (2) to assess the effectiveness of introducing external knowledge in different $N$-way-$M$-shot settings; and (3) to provide empirical evidence that our adaptive knowledge offset can flexibly adjust the impact of the support set and prior knowledge on event prototypes, making the model more accurate and generalizable.




\subsection{Experimental Settings}
We evaluate our method on an aggregated few-shot event detection dataset \emph{FewEvent}\footnote{\url{https://github.com/231sm/Low_Resource_KBP}}~\cite{deng2020meta}. 
\emph{FewEvent} combines two
currently widely-used event detection datasets, the ACE-2005 corpus\footnote{\url{http://projects.ldc.upenn.edu/ace/}} and the TAC-KBP-2017 Event Track Data\footnote{\url{https://tac.nist.gov/2017/KBP/Event/index.html}}, and adds external event types in specific domains including music, film, sports and education~\cite{deng2020meta}.
As a result, \emph{FewEvent} contains 70,852 samples for 19 event types that are further divided into 100 event subtypes. 

In order to match the few-shot settings , we use 88 event types covering a total of 15,681 samples to construct experimental data.
68 event types are selected for training, 10 for validation, and the rest 10 for testing. 
Note that there are no overlapping types between the training, validation and testing sets. 
In order to obtain a convincing result, we conducted 5 random divisions of training and testing for all event types, and the experimental results are averaged as the final result.





The comparisons with our AKE-BML are performed in two aspects, the 
\emph{sample encoder} and the \emph{few-shot learner}. 
We combine different encoders and few-shot learners to obtain different baseline models.
We consider four sample encoders including CNN~\cite{kim2014convolutional}, Bi-LSTM~\cite{huang2015bidirectional}, DMN~\cite{kumar2016ask} and our trigger-attention-based sample encoder TA.
For few-shot learners, we consider Matching Networks (MN)~\cite{vinyals2016matching} and Prototypical Networks (PN)~\cite{snell2017prototypical}. 
We also compare to the SOTA few-shot event detection method DMN-MPN ~\cite{deng2020meta}, which uses a dynamic memory network (DMN) as the sample encoder and a memory-based prototypical network as the few-shot learner.
In addition, in order to verify the effectiveness of our proposed method, we perform an ablation study on our model, which evaluate the model without external knowledge and without dynamic knowledge adaptation.


As a result, the following methods are compared in our experiments:

\begin{itemize}
    \item \textbf{AKE-BML}, our adaptive knowledge-enhanced Bayesian meta-learning method which uses TA encoder as the sample encoder.
    \item \textbf{KB-BML}, a variant of AKE-BML without dynamic knowledge adaption. 
    \item \textbf{TA-BML}, a variant of AKE-BML using our TA encoder but without using external knowledge.
    \item \textbf{DMN-MPN}, dynamic-memory-based prototypical network~\cite{deng2020meta}.
    \item \textbf{Encoder+Learner}, combinations of various sample encoders and event type learners (e.g.\ CNN+MN and TA+PN).
\end{itemize}

We use stochastic gradient descent \cite{bottou2012stochastic} as the optimizer in training with the learning rate $1 \times 10^{-5}$. 
The sampling times $N_s$ of Monte Carlo sampling and update step size $\epsilon$ are set to 10 and 0.01 respectively. 
The update times of stochastic gradient Langevin dynamics $M$ is set to 5. 
We use dropout after the sample encoder and the knowledge encoder to avoid over-fitting; the dropout rate is set to 0.5.
We evaluate the performance of event detection with $F_1$ and $Accuracy$ scores.


\subsection{Main Results}
\begin{table*}[h!t]\large
\centering
    \resizebox{0.9\textwidth}{!}{
    \begin{tabular}{l|ccc|ccc}
    \toprule[0.75pt]
    \multirow{2}{*}{Model} & \multicolumn{1}{c}{\begin{tabular}[c]{@{}c@{}} 5-Way-5-Shot \end{tabular}} & \multicolumn{1}{c}{\begin{tabular}[c]{@{}c@{}}5-Way-10-Shot\end{tabular}} & \multicolumn{1}{c|}{\begin{tabular}[c]{@{}c@{}}5-Way-15-Shot\end{tabular}} & \multicolumn{1}{c}{\begin{tabular}[c]{@{}c@{}}10-Way-5-Shot\end{tabular}} & \multicolumn{1}{c}{\begin{tabular}[c]{@{}c@{}}10-Way-10-Shot\end{tabular}} & \multicolumn{1}{c}{\begin{tabular}[c]{@{}c@{}}10-Way-15-Shot\end{tabular}} \\
     
     & $F_1$/$Accuracy$   & $F_1$/$Accuracy$   & $F_1$/$Accuracy$    & $F_1$/$Accuracy$     & $F_1$/$Accuracy$     & $F_1$/$Accuracy$            \\ 
     \midrule[0.75pt]
     Bi-LSTM+MN$^\S$  & 58.19/58.48     & 61.26/61.45     & 65.55/66.04     & 46.43/47.62    & 51.97/52.60     & 56.27/56.47 \\ 
     CNN+MN$^\S$      & 59.30/60.04     & 64.81/65.15     & 68.35/68.58     & 44.85/45.80    & 50.14/50.67     & 54.13/54.49 \\ 
     DMN+MN$^\S$      & 66.09/67.18     & 68.92/69.33     & 70.88/71.17     & 52.81/54.12    & 58.04/58.38     & 61.63/62.01 \\
     TA+MN       & 66.83/67.55     & 69.12/69.64     & 71.13/71.59     & 53.49/55.47    & 59.58/60.01     & 62.41/63.11 \\
     \midrule[0.5pt]
     Bi-LSTM+PN$^\S$  & 62.42/62.72     & 64.65/64.71     & 68.23/68.39     & 53.15/53.59    & 55.87/56.19     & 60.34/60.87 \\
     CNN+PN$^\S$      & 63.69/64.89     & 69.64/69.74     & 70.42/70.52     & 51.12/51.51    & 53.80/54.01     & 57.89/58.28 \\
     DMN+PN$^\S$      & 72.08/72.43     & 72.47/73.38     & 73.91/74.68     & 59.95/60.07    & 61.48/62.13     & 65.84/66.31 \\
     TA+PN       & 73.66/73.92     & 73.81/74.63     & 75.69/76.31     & 61.25/61.88    & 63.89/64.31     & 66.21/67.59 \\
     \midrule[0.5pt]
     DMN-MPN$^\S$     & 73.59/73.86     & 73.99/74.82     & 76.03/76.57     & 60.98/62.44    & 63.69/64.43     & 67.84/68.35 \\
     \midrule[0.5pt]
     TA-BML      & 73.37/73.59     & 74.02/74.63     & 75.52/75.83    & 61.43/62.59    & 63.28/63.96     & 66.27/67.49  \\
     KB-BML      & 74.63/75.07     & 75.06/75.63     & 80.69/81.12     & 65.99/66.82    & 67.47/68.08     & 73.89/74.06 \\
     AKE-BML      & \textbf{88.99}/\textbf{89.36}     & \textbf{90.10}/\textbf{91.48}     & \textbf{91.40}/\textbf{92.34}     & \textbf{84.55}/\textbf{84.94}    & \textbf{86.03}/\textbf{87.73}     & \textbf{87.13}/\textbf{87.45}              \\ 
    \bottomrule[0.75pt]
    \end{tabular}
    }
    \caption{Accuracy and $F_1$ scores of all compared methods. $^\S$ denotes the results that are directly taken from the original paper~\cite{deng2020meta}, due to the unavailability of the source code. 
    }
    \label{tab:overall}
    \vspace{-3mm}
    \end{table*}

As shown in Table \ref{tab:overall}, we compare methods on $F_1$ and $Accuracy$ scores. 
We observe the followings:
\begin{itemize}
\item Our full model AKE-BML outperforms all other methods on both $Accuracy$ and $F_1$ scores across all settings.
Compared with the SOTA method DMN-MPN, AKE-BML achieves a substantial improvement of 15--23 absolute $F_1$ points in all $N$-way-$M$-shot settings. 
It shows  our adaptive knowledge-enhanced Bayesian meta-learning method can effectively utilize external knowledge and adjust it according to the support set,  thus build better prototypes of event types. Please see Appendix~\ref{sec:m-shot} and ~\ref{n-way} for a detailed performance analysis over various $N$-way and $M$-shot settings.
\item With the sample encoders (Bi-LSTM, CNN, DMN and TA) fixed, it can be observed that prototypical networks (PN) consistently outperforms matching networks (MN). 
DMN-MPN performs better than PN-based methods, because the dynamic memory network can extract key information from the support set through multiple iterations. However, DMN-MPN only considers the information of a few samples in each support set, hence suffering from insufficient sample diversity similar to PN- and MN-based methods. 

\item 
TA-BML performs similarly with DMN-MPN under the settings of $N$-way-$5$-shot and $N$-way-10-shot, but slightly worse under the $N$-way-15-shot setting.
One possible explanation is that when the number of samples in the support set is larger, MPN can generate higher-quality prototypes.
In addition, the performance of TA-BML is not as good as KB-BML, which shows the importance of introducing external knowledge.

\item 
Compared with KB-BML, our full model AKE-BML can effectively solve the problem of deviation between knowledge and event types, and generate event prototypes with better generalization through knowledge.
Compared with TA-BML, which does not incorporate external knowledge, AKE-BML achieves an even larger performance advantage, which further demonstrates the effectiveness of external knowledge. 

\end{itemize}

\subsection{Case Study}
We present a case study on the dynamic knowledge adaptation between the support set and the corresponding event knowledge to demonstrate our model's ability to learn robust event prototypes.

\begin{table*}[h]
    \centering
    \includegraphics[trim={0.3cm 0.7cm 0.8cm 0.7cm},clip,width=0.85\textwidth]{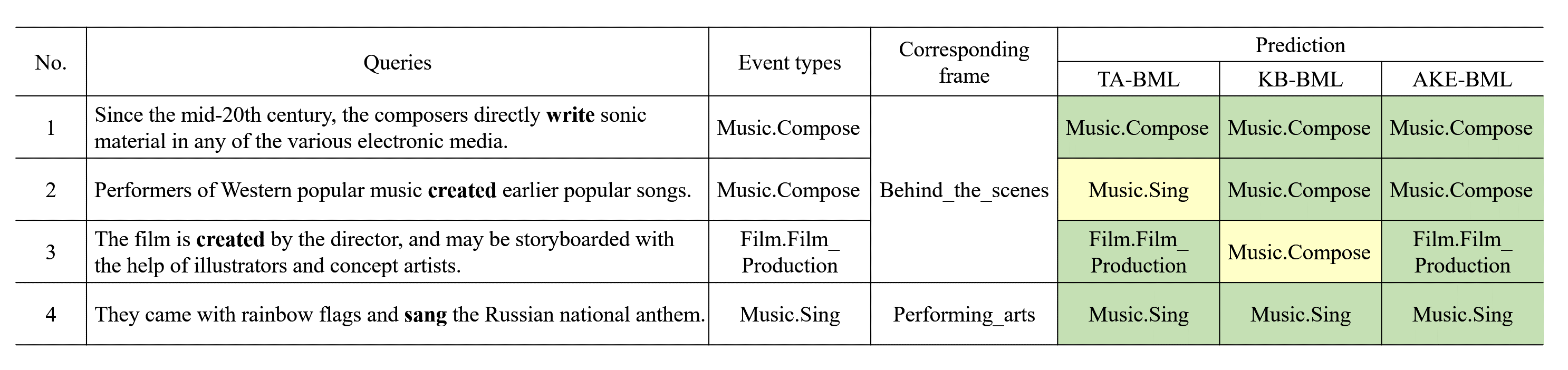}
    \caption{A case study of three event types. Words in bold indicate candidate trigger words.}
    \label{fig:case}
    \vspace{-4mm}
\end{table*}

\begin{table*}[!h]
    \centering

    \includegraphics[trim={0.1cm 0 0.1cm 0},clip,width=0.85\textwidth]{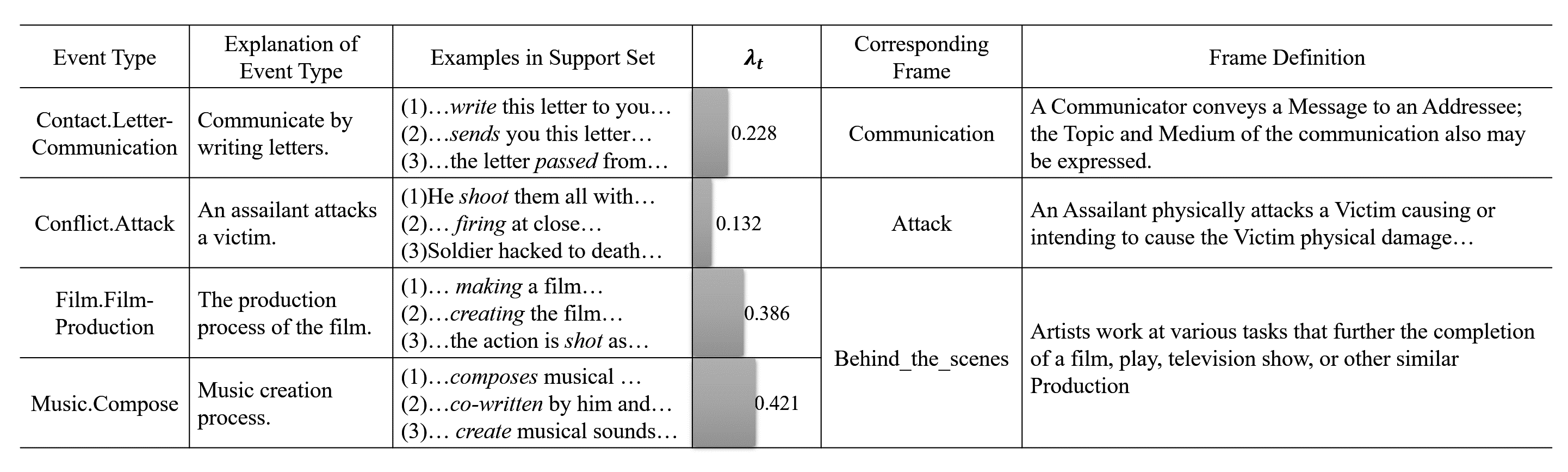}
    \caption{$\lambda_t$ values corresponding to different event types. The larger the $\lambda_t$, the greater the dependence of the prototype vector on the support set.}
    \label{fig:lamb}
    \vspace{-4mm}
\end{table*}

\subsubsection{Predictions for Specific Cases} 
We select three event types as target categories to illustrate the contributions of each main component of our model. 
The event types are \emph{Music.Compose}, \emph{Music.Sing} and \emph{Film.Film\_Productution}.
The sample contexts of \emph{Music.Compose} and \emph{Music.Sing} are similar, while \emph{Music.Compose} and \emph{Film.Film\_Productution} share the same frame, which is \emph{Behind\_the\_scenes}. 

As shown in Table~\ref{fig:case}, only AKE-BML correctly predicts on all samples.
TA-BML, the model without introducing knowledge, wrongly predicts the second sample of \emph{Music.Compose} to be \emph{Music.Sing}, due to their similar contexts.
By introducing knowledge, both KB-BML and AKE-BML avoid this error, indicating that external knowledge can enrich event information based on the support set.
For KB-BML, as \emph{Music.Compose} and \emph{Film.Film\_Productution} share the same superordinate frame, the prototype of \emph{Music.Compose} cannot distinguish between \emph{Music.Compose} and \emph{Film.Film\_Productution}, so it wrongly classifies the third sample as \emph{Music.Compose}.
With our adaptive knowledge offset, AKE-BML can deal with sample similarity and knowledge deviation issues at the same time, thus it correctly classifies all samples. 


\begin{figure}[t]
    \centering
    \includegraphics[trim={1.1cm 0.8cm .9cm 1.1cm},clip,width=0.40\textwidth]{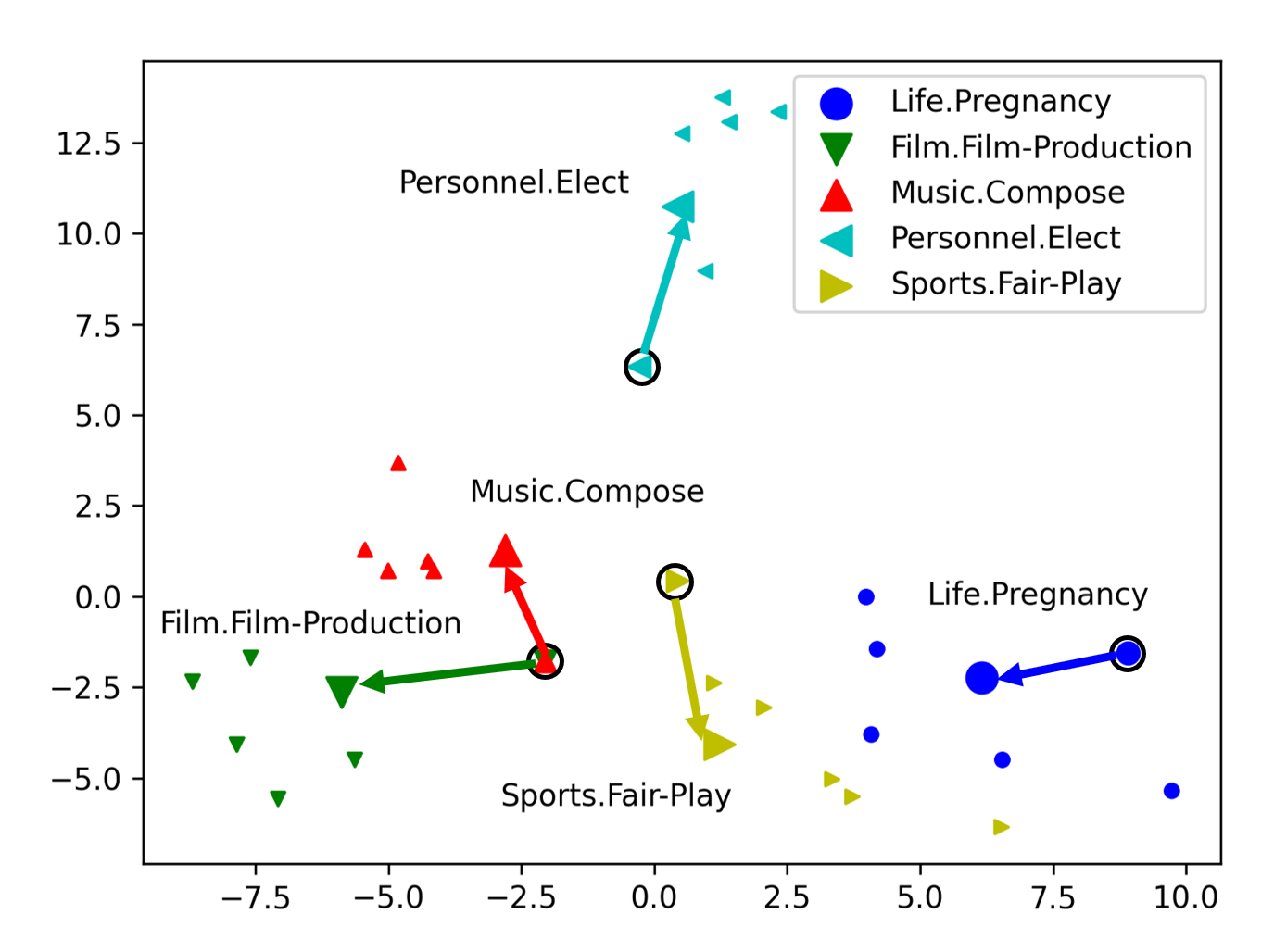}
    \caption{Visualization of event prototypes, prior knowledge, and event samples learned by AKE-BML in the 5-way-5-shot setting. The large, solid shapes denote event prototypes, the large shapes with circle outlines denote the prior knowledge, and the small shapes denote samples. Samples are marked by the color of their corresponding event types. The arrows indicate the adaption of prior knowledge to the prototype. Note that \emph{Music.Compose} and \emph{Film.Film-Production} share the same frame \emph{Behind\_the\_scenes}.}
    \label{fig:visual}
     \vspace{-5mm}
\end{figure}
\subsubsection{Visualization of Prototypes} 
We use Latent Dirichlet Allocation (LDA)~\cite{blei2003latent} to reduce the dimensionality of the prototypes, sample encodings and prior knowledge encodings. Figure~\ref{fig:visual} visualizes five event type prototypes (large solid shapes), their aligned frames (large solid shapes with circle outlines) in FrameNet and some corresponding samples (small solid shapes). 
Each event type and its samples are coded with the same color. 

In general, the samples and prototypes belonging to one event type are close in the space and different event types are far away from each other.
Prior knowledge is distributed in different places in the space, which roughly determines the distribution of event prototypes. For example, the samples of \emph{Life.Pregnancy} and \emph{Sports.Fair-Play} are close to their respective event prototypes. 
Meanwhile, the distances between their prior knowledge is large, making their prototypes easily distinguishable.

It can also be seen that the event prototypes are closer to their samples than to the prior knowledge, which reflects the benefits of our proposed learnable knowledge offset.
The visualization demonstrates the effectiveness of introducing external knowledge and our adaptive knowledge offset's ability to balance the impact of the support set and prior knowledge on the event prototypes.

\subsubsection{$\lambda_t$ of Different Event Types}
As shown in Formula~(\ref{eq:delt}), we use the learnable parameter $\lambda_t$ to generate knowledge offsets. 
$\lambda_t$ accounts for the deviation of the prior knowledge (i.e.\ a frame) from the event type it represents, and adaptively corrects this deviation using information of the support set.
When the frame corresponding to the event type accurately expresses its semantics, the $\lambda_t$ value should be small.
When the knowledge is the super-ordinate frame of the event type (i.e., the frame cannot accurately describe the event semantics), the $\lambda_t$ value should be large, so that the support set can be used to modify the prior knowledge to ensure that the prototype precisely represents the current event type.

Table~\ref{fig:lamb} shows four different event types, their corresponding frames and $\lambda_t$ values.
The $\lambda_t$ of \emph{Conflict.Attack} is a small value 0.132, as the event type \emph{Conflict.Attack} closely matches the frame \emph{Attack}.
The event type \emph{Contact.Letter-Communication} matches the frame \emph{Communication}. \emph{Communication} does not contain the semantics of "by writing letters", but the core semantics is the same as \emph{Contact.Letter-Communication}. Therefore, $\lambda_t$ is small, at 0.228, which is still larger than the $\lambda_t$ of \emph{Conflict.Attack}.
The event types \emph{Film.Film-Production} and \emph{Music.Compose} share the same super-ordinate frame \emph{Behind\_the\_scenes} as prior knowledge, but the semantics of \emph{Behind\_the\_scenes} is too abstract for \emph{Film.Film-Production} and \emph{Music.Compose}.
Thus, it can be seen that the $\lambda_t$ values corresponding to these two event types are relatively large: 0.386 for \emph{Film.Film-Production} and  0.421 for \emph{Music.Compose}.

The above cases demonstrate that our model is able to balance the influence of the support set and the knowledge on event prototypes through $\lambda_t$, and consequentially obtain highly accurate and generalizable prototypes.

\section{Conclusion}
In this paper, we proposed an Adaptive Knowledge-enhanced Bayesian Meta-Learning (AKE-BML) method for few-shot event detection. We alleviate the insufficient sample diversity problem in few-shot learning by leveraging the external knowledge base \emph{FrameNet} to learn prototype representations for event types. We further tackle the uncertainty and incompleteness issues in knowledge coverage with a novel knowledge adaptation mechanism. 

The comprehensive experimental results demonstrate that our proposed method substantially outperforms state-of-the-art methods, achieving a performance improvement of at least 15 absolute points of $F_1$. 
In the future, we plan to extend our proposed AKE-BML method to the few-shot event extraction task, which considers both event detection and argument extraction. We also plan to explore the zero-shot and incremental event extraction scenarios.

\section{Acknowledgement}
Research in this paper was partially supported by the Fundamental Research Funds for the Central Universities (2242021k10011).

\bibliographystyle{acl_natbib}
\bibliography{anthology,acl2021}


\clearpage
\section*{Appendix}
\appendix

\section{FrameNet}\label{sec:knowledge}

An important problem in the few-shot event detection task is the insufficient diversity of support set samples. 
There are only a few labeled samples in the support set, which results in the model unable to construct high-quality prototype features of event types.
To address this problem, we introduce the \emph{FrameNet}~\cite{baker1998berkeley} as an external knowledge base of event types.
\emph{FrameNet} is a linguistic resource storing information about lexical and predicate-argument semantics.
Each frame in \emph{FrameNet} can be taken as a semantic frame of an event type~\cite{liu2016leveraging}, which can be used as background knowledge for event types to assist event detection~\cite{liu2016leveraging,fillmore2006can}.
Figure \ref{fig:example1} shows an example frame defining \emph{Attack}. 
We can see the arguments involved in an \emph{Attack} event and their roles.
The linguistic units (LUs) of the frame \emph{Attack} are the possible trigger words for the corresponding event.
The frame is an important complementary source of knowledge to the support set.
We match a frame in \emph{FrameNet} to each event type, based on the event name, as its knowledge.
In practice, \emph{FrameNet} does not provide complete coverage of all event types, nor does every event type have an exact frame matched in \emph{FrameNet}.
For event types that cannot be exactly matched, we assign the frame corresponding to their super-ordinate event.
For example, there is no corresponding frame for \emph{Contact.Online-Chat}, so we assign it to the frame \emph{Chatting}, which corresponds to the event type \emph{Contact.Chat}.

    \section{Gradient of posterior distribution}
    \label{A:GPD}
    In order to show the change of the prototype vector after adding the knowledge shift, the gradient $\nabla_{\hat{\textbf{V}}_{T_S}} \log \ p(Y_S|X_S,\hat{\textbf{V}}_{T_S})p(\hat{\textbf{V}}_{T_S}|\mathscr{F})$ in iteration
    \vspace{-2mm} 
    \begin{eqnarray}
    \label{eq:update}
        \hat{\textbf{V}}_{T_S} && \gets \hat{\textbf{V}}_{T_S} + \sqrt{\epsilon} \boldsymbol{z}  \\
        &&+ \frac{\epsilon}{2} \nabla_{\hat{\textbf{V}}_{T_S}} \log \ p(Y_S|X_S,\hat{\textbf{V}}_{T_S})p(\hat{\textbf{V}}_{T_S}|\mathscr{F}), \nonumber 
    \vspace{-2mm} 
\end{eqnarray}
    is expanded. For ease of explanation, we only calculate the gradient of the prototype vector $\hat{\textbf{v}}_{t}$.
    We denote the gradient of the original posterior distribution as $\textbf{g}_{\hat{\textbf{v}}_{t}}^o$, the gradient of the knowledge-shifted posterior distribution as $\textbf{g}_{\hat{\textbf{v}}_{t}}^s$.
    We first calculate  $\textbf{g}_{\hat{\textbf{v}}_{t}}^o$:
    \begin{equation}
        \label{eq:detail_o}
        \begin{aligned}
            \textbf{g}_{\hat{\textbf{v}}_{t}}^o   = & \nabla_{\hat{\textbf{v}}_{t}} \log \ p(Y_S|X_S,\hat{\textbf{V}}_{T_S})p^o(\hat{\textbf{V}}_{T_S}|\mathscr{F}) \\
             =& \nabla_{\hat{\textbf{v}}_{t}} \log \ p(Y_S|X_S,\hat{\textbf{V}}_{T_S}) + \log \ p^o(\hat{\textbf{V}}_{T_S}|\mathscr{F}) \\
             =& \nabla_{\hat{\textbf{v}}_{t}} \log \ p(Y_S|X_S,\hat{\textbf{V}}_{T_S}) + \\ &  \nabla_{\hat{\textbf{v}}_{t}} \log \ p^o(\hat{\textbf{V}}_{T_S}|\mathscr{F}) \\ 
             =& \nabla_{\hat{\textbf{v}}_{t}} \log \prod_{s \in S, y_s = t} p(y_s|x_s,\hat{\textbf{v}}_{t}) + \\ & \nabla_{\hat{\textbf{v}}_{t}} \log \ p^o(\hat{\textbf{v}}_{t}|\mathscr{F}) \\ 
             =& \textbf{g}_{\hat{\textbf{v}}_{t}}^{o,l}  +  \textbf{g}_{\hat{\textbf{v}}_{t}}^{o,p},
        \end{aligned}
    \end{equation}
    where $\textbf{g}_{\hat{\textbf{v}}_{t}}^{o,l} = \nabla_{\hat{\textbf{v}}_{t}} \log \prod_{s \in S, y_s = t} p(y_s|x_s,\hat{\textbf{v}}_{t})$ and $\textbf{g}_{\hat{\textbf{v}}_{t}}^{o,p} = \nabla_{\hat{\textbf{v}}_{t}} \log \ p^o(\hat{\textbf{v}}_{t}|\mathscr{F})$.
    The prior distribution is
    \begin{equation}
        \label{eq:detail_o1}
        \begin{aligned}
            p^o(\hat{\textbf{v}}_{t}|\mathscr{F}) & = \mathcal{N}(\textbf{v}_t|\textbf{h}_t,\boldsymbol{I}) \\
            & =\left( 2 \pi \right)^{- \frac{d}{2}} e^{- \frac{1}{2} \left( \hat{\textbf{v}}_t - \textbf{h}_t \right)^2},
        \end{aligned}
    \end{equation}
    the gradient of the logarithm of prior distribution to $\hat{\textbf{v}}_{t}$ is:

    \begin{equation}
        \label{eq:detail_o1}
        \begin{aligned}
            \textbf{g}_{\hat{\textbf{v}}_{t}}^{o,p} & = \left( \log \left( 2 \pi \right)^{- \frac{d}{2}} \right)  \left( \textbf{h}_t - \hat{\textbf{v}}_t \right) \\
            & = C \left( \textbf{h}_t - \hat{\textbf{v}}_t \right),
        \end{aligned}
    \end{equation}
    where $C= \log \left( 2 \pi \right)^{- \frac{d}{2}} $ is a constant, $d$ is the dimension of prototype. The gradient of the log-likelihood on support set is
    \begin{equation}
        \label{eq:detail_o1}
        \begin{aligned}
            & \log \ \prod_{s \in S, y_s = t} p(y_s|x_s,\hat{\textbf{v}}_{t}) \\& = \log\  \prod_{s \in S, y_s = t} \frac{exp(\bm{\mathcal{E}}(x_s) \cdot \hat{\textbf{v}}_t)}{\sum_{t' \in T_S} exp(\bm{\mathcal{E}}(x_s) \cdot \hat{\textbf{v}}_{t'})} \\
            & = \sum_{s \in S, y_s = t} \log \ \frac{exp(\bm{\mathcal{E}}(x_s) \cdot \hat{\textbf{v}}_t)}{\sum_{t' \in T_S} exp(\bm{\mathcal{E}}(x_s) \cdot \hat{\textbf{v}}_{t'})}.
        \end{aligned}
    \end{equation}

    The gradient of the log-likelihood to $\hat{\textbf{v}}_t$ is
    \begin{equation}
        \label{eq:detail_op}
        \begin{aligned}
            \textbf{g}_{\hat{\textbf{v}}_{t}}^{o,l} =& \nabla_{\hat{\textbf{v}}_{t}} \sum_{s \in S, y_s = t} \log \ \frac{exp \left(\bm{\mathcal{E}}(x_s) \cdot \hat{\textbf{v}}_t \right)}{\sum_{t' \in T_S} exp \left(( \bm{\mathcal{E}}(x_s) \cdot \hat{\textbf{v}}_{t'} \right)} \\
            =& \sum_{s \in S, y_s = t} \nabla_{\hat{\textbf{v}}_{t}} \log \ \frac{exp(\bm{\mathcal{E}}(x_s) \cdot \hat{\textbf{v}}_t)}{\sum_{t' \in T_S} exp(\bm{\mathcal{E}}(x_s) \cdot \hat{\textbf{v}}_{t'})} \\
            =& \sum_{s \in S, y_s = t} \nabla_{\hat{\textbf{v}}_{t}} (\bm{\mathcal{E}}(x_s) \cdot \hat{\textbf{v}}_t)  \\
            & - \nabla_{\hat{\textbf{v}}_{t}} \log \ \sum_{t' \in T_S} exp(\bm{\mathcal{E}}(x_s) \cdot \hat{\textbf{v}}_{t'}) \\ 
            =& \sum_{s \in S, y_s = t} \bm{\mathcal{E}}(x_s) - \frac{\bm{\mathcal{E}}(x_s) exp(\bm{\mathcal{E}}(x_s) \cdot \hat{\textbf{v}}_t) }{\sum_{t' \in T_S} exp(\bm{\mathcal{E}}(x_s) \cdot \hat{\textbf{v}}_{t'})} \\
            =& \sum_{s \in S, y_s = t} (1-p_{s}^{(t)})\bm{\mathcal{E}}(x_s),
        \end{aligned}
    \end{equation}
    where $p_{s}^{(t)}=p(y_s|x_s,\hat{\textbf{v}}_{t})$ is the probability of correct classification of sample $s$ in support set. Then we get
    \begin{equation}
        \begin{aligned}
            \textbf{g}_{\hat{\textbf{v}}_{t}}^o =\sum_{s \in S, y_s = t} (1-p_{s}^{(t)})\bm{\mathcal{E}}(x_s) + C(\textbf{h}_t-\hat{\textbf{v}}_t).
        \end{aligned}
    \end{equation}
    Then we calculate $\textbf{g}_{\hat{\textbf{v}}_{t}}^s$, The only difference between calculating $\textbf{g}_{\hat{\textbf{v}}_{t}}^s$ and $\textbf{g}_{\hat{\textbf{v}}_{t}}^o$ is  $\textbf{g}_{\hat{\textbf{v}}_{t}}^s$ use the knowledge-shifted prior distribution
    \begin{equation}
        \label{eq:detail_o1}
        \begin{aligned}
            p(\hat{\textbf{v}}_{t}|\mathscr{F}) & = \mathcal{N}(\textbf{v}_t|\textbf{h}_t + \Delta \textbf{h}_t,\boldsymbol{I}) \\
            & =\left( 2 \pi \right)^{- \frac{d}{2}} e^{- \frac{1}{2} \left( \hat{\textbf{v}}_t - \textbf{h}_t - \Delta \textbf{h}_t \right)^2}.
        \end{aligned}
    \end{equation}

    Same as the original posterior gradient, we have
    \begin{equation}
        \label{eq:detail_o}
        \begin{aligned}
            \textbf{g}_{\hat{\textbf{v}}_{t}}^s  =&  \nabla_{\hat{\textbf{v}}_{t}} \log \prod_{s \in S, y_s = t} p(y_s|x_s,\hat{\textbf{v}}_{t}) \\
            &+ \nabla_{\hat{\textbf{v}}_{t}} \log \ p(\hat{\textbf{v}}_{t}|\mathscr{F}) \\ 
            =& \textbf{g}_{\hat{\textbf{v}}_{t}}^{s,l} + \textbf{g}_{\hat{\textbf{v}}_{t}}^{s,p},
        \end{aligned}
    \end{equation}
    where $\textbf{g}_{\hat{\textbf{v}}_{t}}^{s,l} = \textbf{g}_{\hat{\textbf{v}}_{t}}^{o,l} = \sum_{s \in S, y_s = t} (1-p_{s}^{(t)})\bm{\mathcal{E}}(x_s)$. The gradient of the logarithm of knowledge-shifted prior distribution to $\hat{\textbf{v}}_{t}$ is:
    \begin{equation}
        \label{eq:detail_o1}
        \begin{aligned}
            \textbf{g}_{\hat{\textbf{v}}_{t}}^{s,p} & = \left( \log \left( 2 \pi \right)^{- \frac{d}{2}} \right)  \left( \textbf{h}_t + \Delta \textbf{h}_t - \hat{\textbf{v}}_t \right) \\
            & = C \left( \textbf{h}_t + \bm{\lambda}_t \odot (\textbf{m}_t - \textbf{h}_t) - \hat{\textbf{v}}_t \right) \\ 
            & = C \left( (\mathds{1}-\bm{\lambda}_t) \odot \textbf{h}_t  - \hat{\textbf{v}}_t \right) + C \bm{\lambda}_t \odot \textbf{m}_t,
        \end{aligned}
    \end{equation}
    where $\mathds{1}$ is a $|\textbf{h}_t|$-dimensional vector, and each element of $\mathds{1}$ is 1. then we get

    \begin{equation}
        \label{eq:detail_o}
        \begin{aligned}
            \textbf{g}_{\hat{\textbf{v}}_{t}}^s   =&  \textbf{g}_{\hat{\textbf{v}}_{t}}^{s,l} + \textbf{g}_{\hat{\textbf{v}}_{t}}^{s,p} \\
            =& \sum_{s \in S, y_s = t} (1-p_{s}^{(t)})\bm{\mathcal{E}}(x_s) \\
            &+ C \left( (\mathds{1}-\bm{\lambda}_t) \odot \textbf{h}_t  - \hat{\textbf{v}}_t \right) \\
            &+ C \bm{\lambda}_t \odot \textbf{m}_t,
        \end{aligned}
    \end{equation}
    bring $\textbf{m}_t=\frac{1}{M} \sum_{s \in S, y_s = t} \bm{\mathcal{E}}(x_s)$ into the above formula, we get
    \begin{equation}
        \label{eq:detail_o}
        \begin{aligned}
            \textbf{g}_{\hat{\textbf{v}}_{t}}^s  = &\sum_{s \in S, y_s = t} \left[ (1 - p_{s}^{(t)}) \bm{\mathcal{E}}(x_s) + \frac{C \bm{\lambda}_t}{M} \odot \bm{\mathcal{E}}(x_s) \right] \\ & + C \left( (\mathds{1}-\bm{\lambda}_t) \odot \textbf{h}_t  - \hat{\textbf{v}}_t \right).
        \end{aligned}
    \end{equation}
    
Note that, when knowledge adaption is not used, the form of the prior knowledge distribution of the prototype is as follows,
\begin{equation}
    \label{eq:priorfixed}
       p^o(\textbf{V}_{T_S} |\mathscr{F}) = \prod_{t \in T_S} p^o(\textbf{v}_t|\textbf{h}_t) = \prod_{t \in T_S} \mathcal{N}(\textbf{v}_t|\textbf{h}_t,\boldsymbol{I}).
\end{equation}

To intuitively show the influence of the knowledge-adapted posterior distribution on the prototype vector, we expand the gradient $\nabla_{\hat{\textbf{V}}_{T_S}} \log \ p(Y_S|X_S,\hat{\textbf{V}}_{T_S})p(\hat{\textbf{V}}_{T_S}|\mathscr{F})$ in Eqn \ref{eq:update}. 
For ease of explanation,  we only calculate the gradient of the prototype vector $\hat{\textbf{v}}_{t}$. 
Denote the gradient of the original posterior distribution without knowledge adaption as $\textbf{g}_{\hat{\textbf{v}}_{t}}^o$, 

\begin{equation}
\label{eq:go}
    \textbf{g}_{\hat{\textbf{v}}_{t}}^o = \sum_{s \in S, y_s = t} (1-p_{s}^{(t)})\bm{\mathcal{E}}(x_s) + C(\textbf{h}_t-\hat{\textbf{v}}_t),
\end{equation}

and the gradient of the knowledge-adapted posterior distribution as $\textbf{g}_{\hat{\textbf{v}}_{t}}^s$ from Eqn~\ref{eq:detail_o}.

%
 Comparing Eqn~\ref{eq:detail_o} and Eqn~\ref{eq:go}, it can be seen that the posterior distribution without knowledge adaption cannot dynamically balance the influence of the knowledge and the support set on the prototype vector, whereas the knowledge-adapted posterior distribution can adjust their contributions to the prototype vector through $\bm{\lambda}_t$. The parameters in Eqn~\ref{eq:lambda} will be updated by the log-likelihood on the query set. This allows the model to reasonably choose the weight of the knowledge and the support set, and obtain prototype vectors with better generalization.

\section{$M$-shot Evaluation}\label{sec:m-shot}
In this section, we illustrate the effectiveness of adaptive knowledge-enhanced Bayesian meta-learning under different $M$-shot settings, such as $N$-way-$5$-shot, $N$-way-$10$-shot and $N$-way-$15$-shot. 
As shown in Table \ref{tab:overall} in the main paper, as $M$ increases, the performance of all models improves, which shows that increasing the number of samples in the support set can provide more pertinent event type-related features.
At the same time, it can be seen that from 15-shot to 5-shot, the previous methods suffer a significantly larger performance degradation than AKE-BML.
This observation shows our model's strong robustness against low sample diversity due to the incorporation of external knowledge. 

The performance of KB-BML is close to that of DMN-MPN in the case of $N$-way-$5$-shot and $N$-way-$10$-shot, and the performance is better in the case of $N$-way-$15$-shot. 
This can be attributed to two factors: (1) the introduction of knowledge can improve the generalization of event prototypes; and (2) increasing the number of samples can reduce the impact of the deviation between knowledge and event types.
When the support set is sufficiently large, the samples in the support set can compensate for the deviation between knowledge and event types, and the knowledge can also improve the generalization of the prototype vector.
However, when $M$ is small, the deviation between knowledge and event types will affect the quality of the prototype vectors.

AKE-BML can well balance the effects of samples and knowledge on the event type prototypes. 
It can be seen that when $M$ is small, the performance of AKE-BML does not decline as quickly as other models, which also proves the effectiveness of knowledge in dealing with the problem of insufficient diversity of the support set. 
At the same time, compared with KB-BML, our adaptive knowledge offset can effectively use the information in the support set to correct the knowledge deviation.


\section{$N$-Way Evaluation}\label{n-way}
\begin{figure}[t]
    \centering
    \includegraphics[scale=0.18]{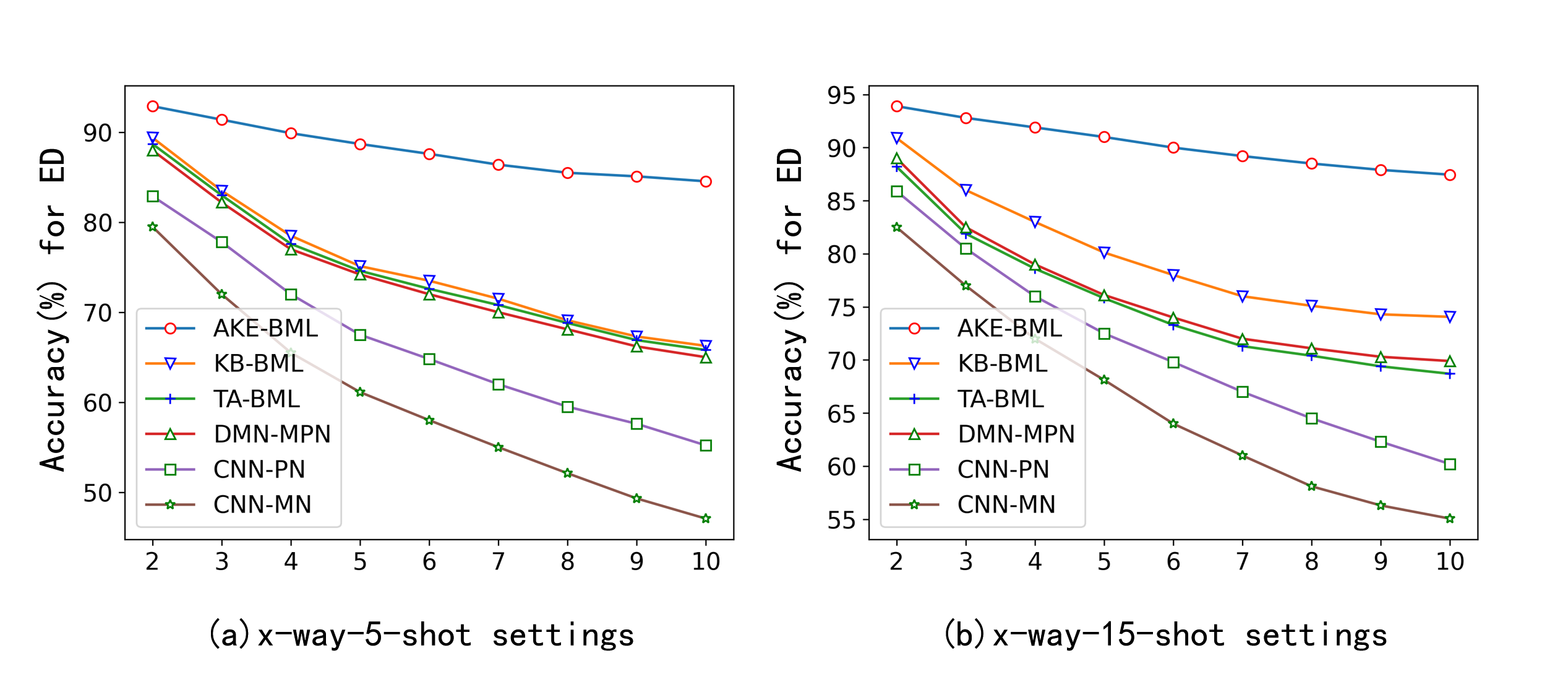}
    \caption{$N$-way ($N=2,\ldots, 10$) evaluation and fixed shot numbers. (a) $N$-way-5-shot. (b) $N$-way-15-shot.}
    \label{fig:xway}
\end{figure}

Figure~\ref{fig:xway} also illustrates model performance with respect to different way values (i.e.\ $N$), while fixing the shot values. 
It can be seen from the figure that when $N$ increases, the performance of previous models decreases faster than AKE-BML, which shows that those models, only relying on the support set, cannot generate more recognizable event prototypes. 
The performance of KB-BML also declines significantly when $N$ increases. 
This is because many event types can only be partially aligned in FrameNet, to its super-ordinate frame, which causes the event prototypes to be indistinguishable to similar event types.

On the contrary, the performance of AKE-BML does not decrease significantly when $N$ increases, which shows that our adaptive knowledge-enhanced Bayesian meta-learning method can enhance the distinguishability of prototype vectors through the learnable knowledge offset. 
These results indicate that our adaptive knowledge-enhanced Bayesian meta-learning is more robust to the changes in the number of ways.


\end{document}